# Evaluating image matching methods for book cover identification

Rabie Hachemi, Ikram Achar, and Biasi Wiga
LaRATIC Laboratory
National Institute of Telecommunications and ICT (INTTIC)
Oran, Algeria
{rhachemi, iachar, bwiga}@inttic.dz

Mahfoud Sidi Ali Mebarek
LSI Laboratory
University of Science and Technology of Oran – Mohamed Boudiaf (USTO)
Oran, Algeria
mahfoud.sidialimebarek@univ-usto.dz

*Abstract*— Humans are capable of identifying a book only by looking at its cover, but how can computers do the same? In this paper, we explore different feature detectors and matching methods for book cover identification, and compare their performances in terms of both speed and accuracy. This will allow, for example, libraries to develop interactive services based on cover book picture. Only one single image of a cover book needs to be available through a database. Tests have been performed by taking into account different transformations of each book cover image. Encouraging results have been achieved.

*Keywords— book cover identification; matching algorithms; feature detection; image processing*

## I.  Introduction

In the recent years, reading and loaning books are no longer the only services provided by libraries. Instead of resisting the rise of digital devices and Internet connectivity, libraries chose to adapt to technology by developing new ways to interact with books, and that, by providing members and visitors with value added services to enrich their experience [1]. Examples of these services include: reading a summary, watching an interview of the author or even playing a quiz around the book.

For these services to be properly implemented, there need to be a one-to-one identification of the books, in order to allow the application to deliver the right content according to the book. In many cases, libraries make use of QR codes [2], which have the benefit of being quick and reliable, however, their implementation is considered heavy and involves human labor. In fact, after the QR codes have been generated, they need to be printed on paper, cut, and finally sticked to all the books one by one [3].

We present in this paper, an approach to identify a book, using only a picture of its cover. Compared to the QR code based methods, it has the benefit to reduce the human factor (since no printing and sticking is required) while keeping the same accuracy. More than that, this approach allows the reuse of the identification system, for example, a library which has the same book in 3 different annexes, will not have to print and stick 3 QR codes, but will only provide a single picture of the cover, in order to activate the identification on all the other copies of the book.

The present paper is organized as follows. The next section is a rapid overview of related works in the field of book cover recognition. Section 3 describes feature detection and matching techniques that are used in our work. In section 4, we present the experiments and discuss the obtained results, and finally in section 5 we conclude and present some perspectives of our work.

## II.  Book Cover Recognition

Book cover recognition is generally seen as an image recognition problem, thus, commonly leading to the use of feature detection algorithms. Many researchers have tackled this issue by focusing on extracting the title from the cover image. For example, Do et al [4], used a combination between the HOG (Histogram of Oriented Gradients) feature descriptor and color information to extract titles from cover. Whereas, Yang et al. [5] used the MSER (Maximally Stable Extremal Regions) feature detector, combined with OCR (Optical Character Recognition) in order to extract it. Once it is correctly extracted, an online search is operated on Amazon bookstore to identify the book.

Meanwhile, other efforts have seen the light, where the search is performed over a database of book covers and not from an online store. For instance, Ho et al. [6] proposed to use a SIFT algorithm (Scale-Invariant Feature Transform) as a feature detector, assuming that the region of interest is the area in the image where the complete book cover appears. For the classification part, an SVM (Support-Vector Machines) classifier was proposed or a K-means algorithm for a faster retrieval (by grouping data into clusters).

Some efforts have also been made to predict a book's genre form the cover using Convolutional Neural Networks [7], however, even if Deep-Learning based methods are outperforming classic image classification algorithms, it is not shown to be useful for the task at hand, given that data is not available beforehand.

In our work, we compare between the following feature detection algorithms: SIFT, SURF, ORB and A-KAZE, outlining the differences in terms of accuracy and speed, for the task of book cover detection and presenting an original methodology to evaluate these metrics in our context.

III. FEATURE DETECTION AND MATCHING

Feature detection and image matching are two interesting areas in computer vision with many applications in various fields (i.e. object detection, video tracking, 3D modeling, etc.). The aim of feature detection is to identify, extract and describe salient points from image data. Based on these features, image matching algorithms are used to decide whether or no different images are similar or representing common objects. For robustness purpose, extracted features should be invariant to different kinds of transformations such as rotation, noise, scaling and illumination.

Feature detection algorithms date back to the 1980s. At that time, various types of features are proposed, such as line segments [8], edge groups or zones [9]. Especially, the corner descriptor proposed by Harris [10] enjoyed great success. Improvements have been made to bring invariance to rotation [11]. However, these features remain sensitive to other types of changes.

In the last two decades, a variety of other techniques have been proposed to overcome these limitations, e.g., SIFT, SURF, ORB and KAZE [12] [13] [14] [15]. These methods differ each other in accuracy and computational time. Accuracy refers to the capacity of extracted features to describe the image content in an invariant way. Computational time concerns both the feature extraction and the matching stages. Various works in literature have attempted to compare these techniques [16] [17], mainly by considering the two above-mentioned criteria. In this work, we made the choice of some of these techniques, which are known to be fast or enough accurate.

A. *Scale Invariant Feature Transform (SIFT)*

The SIFT feature detector was proposed by Lowe in 2004 [12]. The SIFT algorithm consists of four steps. The first step is to search potential interest points by detecting extrema in the difference of Gaussian pyramid (DoG). In the next step, key points are localized and filtered by discarding low-contrast and edge-response candidate points. Each remaining key point is then associated with an intrinsic orientation based on local image gradient directions. Lastly, local image descriptor is generated for each key point based on image gradient magnitude and orientation.

B. *Speeded Up Robust Features (SURF)*

Partially inspired by the SIFT detector, SURF detector was proposed by Bay et al. in 2006 [13]. As the name suggests, it should work faster than SIFT by filtering the image with a square instead of the difference of Gaussian. The square filter can be easily calculated with the help of the integral image. To find the key points, SURF uses a blob detector based on the Hessian matrix. For orientation assignment, it uses Haar wavelet responses in both horizontal and vertical directions within a circular neighborhood around the point of interest. The radius of the circular neighborhood is proportional to the scale in which the point of interest was detected. The obtained responses are weighted by an adequate Gaussian function. The dominant orientation is estimated by calculating the sum of all responses within a sliding orientation window. For feature description, SURF uses also the wavelet responses. A square region centered on the key point and oriented along its orientation is selected and divided into subregions. For each subregion the wavelet responses are used as SURF feature descriptor.

C. *Oriented FAST and Rotated BRIEF (ORB)*

ORB was conceived at OpenCV labs by Rublee et al. in 2011 [14], in part because of the patenting of SIFT and SURF algorithms. It is based on the FAST key point detector and the BRIEF descriptor, with many modifications.

The FAST algorithm (features from accelerated segment test) is a corner detector [18]. It uses a circle of 16 pixels to classify whether a point is a corner. ORB extracts FAST corners from a multi scale image pyramid to produce scale-invariant features. It also applies Harris measure to find top N points. ORB uses a technique called intensity centroid to assign orientation to each key point. Moments are computed to improve rotation invariance.

The BRIEF descriptor (Binary Robust Independent Elementary Features) is a bit string description of an image patch constructed from a set of binary intensity tests [19]. To achieve invariance to in-plane rotation, ORB computes a rotation matrix using the orientation of patch then the BRIEF descriptor are steered according to the orientation.

D. *Accelerated Kaze (AKAZE)*

Based on the KAZE features [20], Alcantarilla et al. proposed the Accelerated-KAZE (A-KAZE) feature detector and descriptor [15].

The KAZE algorithm has been developed to address the problem of non-respect of natural boundaries when using or approximating the Gaussian scale space of an image, which is the case of SIFT and SURF detectors. For that, it detects and describes 2D features in a nonlinear scale space by means of nonlinear diffusion filtering. Indeed, nonlinear diffusion filtering can extract features while maintaining details and reducing noises.

In order to speed up the nonlinear scale space computation, the A-KAZE detector uses the fast explicit diffusion (FED), embedded in pyramid structure. Then A-KAZE uses the Hessian matrix in nonlinear scale space to detect the feature points. For each feature point, a Modified-Local Difference Binary (M-LDB) descriptor is computed.

E. *Image matching methods*

The simplest way to match two feature sets (extracted from two images respectively) is to take the descriptor of each feature in first set, match it with all feature descriptors in second set, and return the best match using some distance measurement. It usually consists of Euclidian distance for SIFT and SURF features, or Hamming distance for ORB and AKAZE features. The closest set of matched features is returned.

Another alternative is to return the k best matches using k-nearest neighbor (k-NN) algorithm. It is assumed that a small NNDR (nearest neighbor distance ratio) between distances to the nearest and the 2nd nearest neighbors results in a good match [12].

In high dimensional spaces, the nearest neighbor matching is time consuming. Muja et al. thus proposed a system to automatically determine the best and fastest algorithm and parameter values that approximate nearest neighbor algorithm given a dataset [21]. They also proposed an algorithm for approximate matching of binary features [22]. A library named FLANN (Fast Library for Approximate Nearest Neighbors) implementing these solutions is available [23].

## IV. EXPERIMENTAL RESULTS

In this section we investigate the feature detection methods described in the previous section, namely SIFT, SURF, ORB and AKAZE, using the different matching techniques. The main purpose is to allow developing interactive services for libraries, responding as well as possible in real time manner. For this reason, we have to find a trade-off between speed and accuracy in performing one-to-one identification of a book.

SIFT and SURF are chosen for their robustness reputation, and because they have become references in many applications in computer vision. ORB and AKAZE also showed good performances and offer viable alternatives to SIFT and SURF. It should be noted that ORB and AKAZE are patent-free.

For our tests, we used a dataset manually created and prepared by ourselves, containing 1400 cover images of 200 books. Each book is represented by a reference cover image and a set of images, according to six different transformations: two rotated images (45 and 90 degree), a cropped image, and three other images captured under changes in illumination, scale and view point. These transformations were defined in such a way that they represent real use case scenarios. The transformed images form a test set, and have to be matched with the 200 reference images. An example of matching two similar cover book images using the different feature detectors is illustrated in *"Fig. 1"*.

First of all, we compared the detection methods by calculating the number of extracted keypoints as well as their computational time. As shown in "TABLE I.", ORB is clearly the fastest method with a minimum of keypoints, which should lead to a fast matching. One can see in *"Fig. 2"* that in the case of ORB and AKAZE, keypoints are concentrated near boundaries, and unlike SIFT and SURF, homogenous regions are almost empty.

TABLE I.  KEYPOINTS EXTRACTION : NUMBER VS TIME

|  | SIFT | SURF | ORB | AKAZE |
|---|---|---|---|---|
| Number of keypoints | 1800 | 9800 | 500 | 3700 |
| Time (sec) | 7.43 | 4.37 | 0.24 | 5.69 |

a. Calculations are performed on images of size 4128x3096
b. The results shown are averaged over many tests

Except ORB, it seems that the other methods are difficult to be used in real time book cover identification application. Indeed, the step of extraction of keypoints from a request image is resource limited since it should be done locally (on the client side), contrary to the matching step which can be achieved with resource adjustments like using GPU (on the server side).

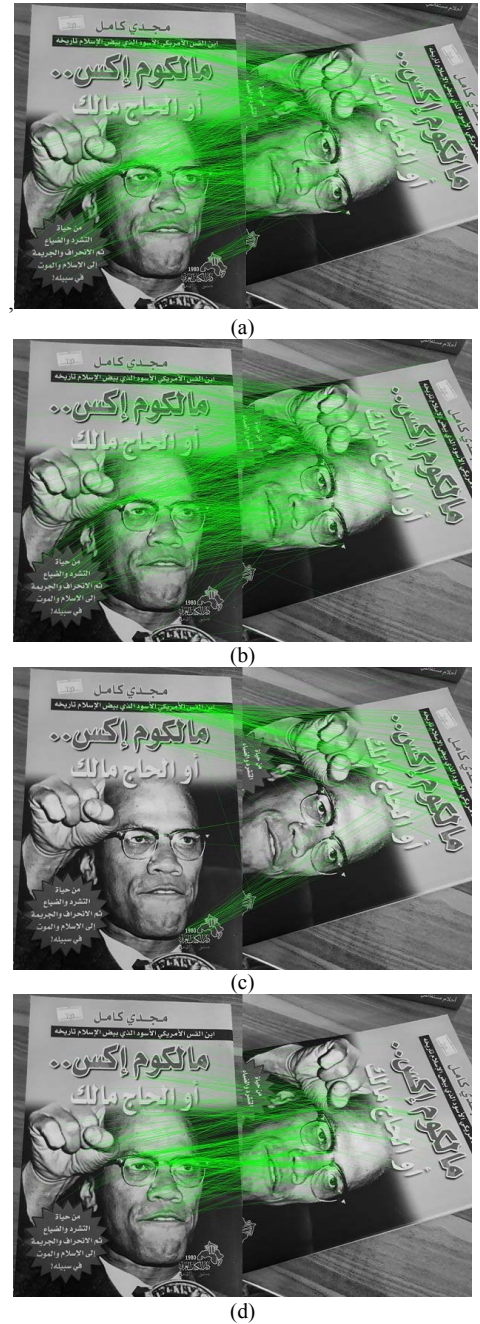

Fig. 1.  Image matching using (a) SIFT (b) SURF (c) ORB (d) AKAZE

At the matching stage, we first used a simple matching (one-nearest neighbor method), i.e. each descriptor from an image is matched with the nearest one in the other image according to a distance measurement (Euclidian distance for SIFT and SURF, and Hamming distance for ORB and AKAZE). The estimated matching times are summarized in

"TABLE II.". As expected, ORB is the fastest method in the matching stage.

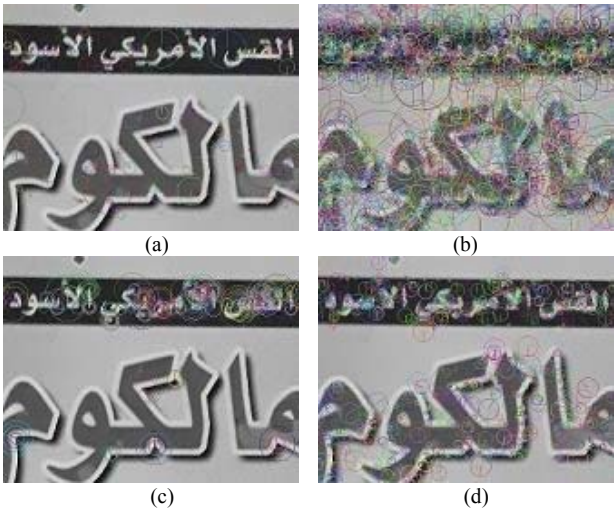

(a) (b) (c) (d)

Fig. 2. Keypoints extracted using (a) SIFT (b) SURF (c) ORB (d) AKAZE

TABLE II. TIME OF MATCHING WITH ONE NEAREST NEIGHBOR METHOD

|  | SIFT | SURF | ORB | AKAZE |
|---|---|---|---|---|
| **Time (sec)** | 0.14 | 2.72 | 0.004 | 0.37 |

In *"Fig. 3"* to *"Fig. 6"*, each curve represents the matching of a test image (a transformed book cover image) with all book cover reference images using SIFT, SURF, ORB and AKAZE descriptors. In this example, it can be seen that there is discrimination between the matching with the target reference image and the matching with other reference images. Thus, in the case of a simple matching, we used the sum of all distances between the pairs of matched descriptors as a discrimination score. This corresponds graphically to the area under curve.

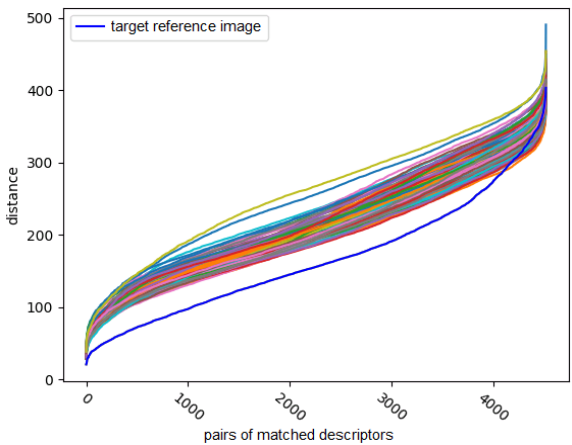

Fig. 3. Example of matching a transformed image with all reference images using SIFT descriptors

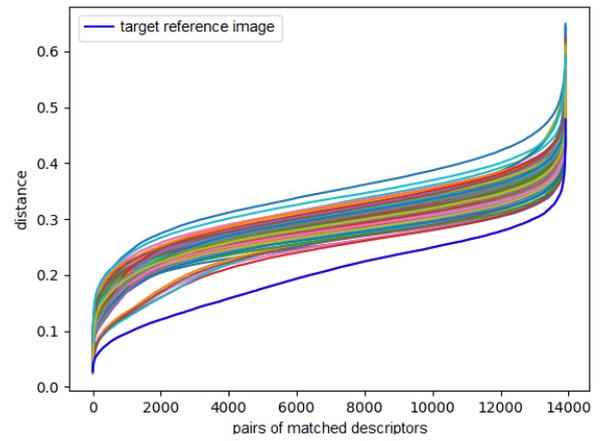

Fig. 4. Example of matching a transformed image with all reference images using SURF descriptors

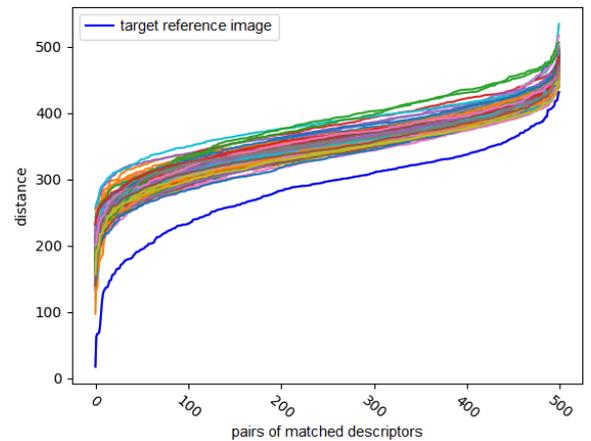

Fig. 5. Example of matching a transformed image with all reference images using ORB descriptors

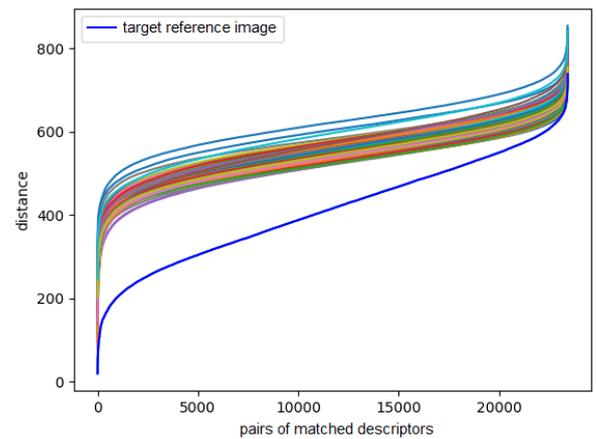

Fig. 6. Example of matching a transformed image with all reference images using AKAZE descriptors

For more robust matching, we tested a kNN matching with k=2 using the FLANN library. The estimating matching times are summarized in "TABLE III.". One again, ORB outperform dramatically the other methods in matching time.

TABLE III.    TIME OF MATCHING WITH K-NN (K=2) USING FLANN

|  | SIFT | SURF | ORB | AKAZE |
|---|---|---|---|---|
| **Time (sec)** | 4.21 | 7.13 | 0.006 | 4.35 |

In this case of matching, the NNDR ratio (Nearest Neighbor Distance Ratio) allows discarding matched descriptors considered irrelevant according to some threshold value. Thus, we used the percentage of retained matched descriptors (matching rate) as a discrimination score. To adjust the threshold value, we varied it and calculated the matching rate of similar images and dissimilar images. A high threshold value (close to 1) leads to weak discrimination while a small threshold value leads to low matching rates even for similar images. It seems that a NNDR between 0.6 and 0.75 is a good choice.

"TABLE IV." presents accuracies when identifying the transformed images (the test set) using the two matching methods. The best accuracies have been achieved by AKAZE. SIFT and SURF provided comparable results. Nevertheless, when taking into account time constraints, ORB led to a relatively good accuracy with simple matching when remaining very fast in both extraction and matching steps. Besides, we have noticed that the mis-identified images in the case of ORB relate to book covers containing only texts. For book covers including paintings or drawings (like children books), ORB performed much better.

TABLE IV.    IDENTICATION ACCURACY

|  | SIFT | SURF | ORB | AKAZE |
|---|---|---|---|---|
| **Simple matching** | 86.4 % | 93.5 % | 84.6 % | 96.3 % |
| **Knn (n=2) matching** | 94.8 % | 88.0 % | 62.2 % | 96.8 % |

## V. CONCLUSION

In this paper, we compared four feature detection methods for the purpose of book cover identification. We chose two patented (SIFT and SURF) and two non-patented (ORB and AKAZE) methods, known to perform well in terms of accuracy or computational time. To allow libraries to develop interactive services based on book cover pictures, we had to find a trade-off between speed and identification accuracy, sufficient enough to discriminate between a similar transformed image and a dissimilar one. In both extraction and matching steps, ORB outperformed significantly the other methods in computational time. The other methods showed good identification accuracies but are time consuming compared to ORB. We have shown that the use of ORB together with a one-nearest neighbor matching yields encouraging results, especially for images rich in details.

In order to improve the identification accuracy, we are looking at the way of integrating other features, especially color features which are relatively simple to calculate. Another interesting prospect could be the clustering of the database images according to the same matching technique (task that can be carried out off-line). The goal is to reduce the number of one-to-one image matching in exploring a database. Tests on large databases need to be made.